\documentstyle[epsf]{llncs} 
\begin{document}

\thispagestyle{empty}

\title{A Semantics-based Communication System for Dysphasic
Subjects\thanks{This work has taken place in the frame of the PVI
({\em Proth\`{e}se Vocale Intelligente}) project, funded by AGEFIPH
and Thomson-CSF. It has involved constant cooperation of the
Rehabilitation Centre of Kerpape (Lorient, Brittany, France).}}

\author{Pascal Vaillant}

\institute{Thomson-CSF/LCR, Computer Science Group,\\
           Domaine de Corbeville, 91404 {\sc Orsay cedex}, {\sc
France}\\
           Phone: (+33) 1 69 33 93 25, Fax: (+33) 1 69 33 08 65\\
           E-mail: \verb+vaillant@lcr.thomson.fr+}

\maketitle

\begin{abstract} Dysphasic subjects do not have complete linguistic
abilities and only produce a weakly structured, topicalized
language. They are offered artificial symbolic languages to help them
communicate in a way more adapted to their linguistic abilities. After
a structural analysis of a corpus of utterances from children with
cerebral palsy, we define a semantic lexicon for such a symbolic
language. We use it as the basis of a semantic analysis process able
to retrieve an interpretation of the utterances. This semantic
analyser is currently used in an application designed to convert
iconic languages into natural language; it might find other uses in
the field of language rehabilitation.  \end{abstract}

\section{Introduction}

The field of Assisted Communication for speech impaired people now
offers a wide range of material or logical devices that produce
audible sentences for the user. Few systems, though, provide a good
communication for subjects whose language abilities, and not only
speech ones, are impaired.

We have tried to tackle the problem of understanding asyntactic
utterances produced by speech and language impaired people through a
technique of semantic analysis. This principle has been implemented in
a computer application, PVI ({\em Proth\`{e}se Vocale Intelligente}),
allowing users to communicate through sequences of icons translated
into French sentences. The same principle had already inspired the
{\sc Compansion} system~\cite{kn:demasco-mccoy}, which converts, with
different AI techniques, sequences of uninflected words into English
sentences.

In this paper, we will expose in a first part what are the language
disabilities we have to cope with, situate them in the frame of
language disorders, and see what type of discourse disorganization
they produce by examining a corpus.

In a second part, we propose a specific technique of semantic analysis
able to analyse this type of discourse. We make some hypotheses on the
structure of the language, draw a model able to represent it, and then
expose the operations one can perform on this model.

We briefly describe the application in which the technique has been
implemented. The application itself is described in more detail
in~\cite{kn:vaillant-checler}.

We finally give some elements of evaluation of the system, as they
emerge both from quantitative (benchmarking) and qualitative (on-site)
evaluation.

\section{Consequences of Dysphasia on the Language}

\subsection{Context of our Study}

Speech and language disorders among children can be so miscellaneous
in their nature as a simple language acquisition delay or a severe and
permanent language deficit.

\begin{enumerate}

\item Language acquisition delay may be correlated with some types of
mental retardation, or in some cases with social or psychological
troubles. Children in this case present some symptoms like lack of
phonological or syntactical control, appearing mainly though as
mistakes of the same nature as typical childhood language mistakes ---
not as a systematic deviant linguistic behaviour.

\item A permanent language deficit may be a consequence of:

\begin{enumerate}

\item a general developmental trouble like {\em autism};

\item a cerebral lesion acquired during childhood --- in this case the
language disorder is referred to as {\em acquired aphasia};

\item a specific language development disorder: {\em dysphasia}.

\end{enumerate}

\end{enumerate}

The subjects we are working with, children with cerebral palsy, suffer
from a global language deficit due to stable cerebral lesions,
consequence of a pre- or perinatal accident (e.g. prolonged anoxia).

It has been shown~\cite{kn:gerard-et-al} that these children present
language disorders which are very close to those of developmental
dysphasia. In a clinical perspective, the diagnostic methods are the
same, and the rehabilitation guidelines are the same in respect to the
proper linguistic troubles. That is why we will further use the term
of dysphasia as a set of clinical symptoms, which can be used to
characterize the subjects in our study.

The techniques described in this paper have been implemented in
communication help software for these children. We will present the
types of speech and language troubles observed among the subjects, as
these troubles may externally appear.

\subsection{Nature of the Language Troubles}

The subjects present various symptoms of speech disability, that may
be classified roughly into two main categories:

\begin{enumerate}

\item speech troubles: phonatory, or articulatory, they hinder the
utterance of speech strictly speaking;

\item language troubles: they show themselves in the use of language
as manipulation of linguistic signs.

\end{enumerate}

Speech troubles occur at different levels as a consequence of the
neuromotor troubles characteristic of cerebral palsy. They can be of
phonatory nature (impossibility to form proper sounds in the oral
cavity: {\em dyslalia}), and of articulatory nature (lack of control
of the muscles which govern articulation: {\em dysarthria}).

Language troubles of the subjects possibly affect many linguistic
competences. They are symptomatically similar to those observed for
dysphasic subjects. We may distinguish:

\begin{enumerate}

\item {\bf Semantic troubles}

They often affect the emergence of abstract concepts and
categories. Some subjects are unable to group into a single concept
several instances of a category. Some may form improper categories,
for example confuse concepts belonging to the same semantic domain.

More scarcely, one may observe, like in adult aphasia, troubles of
lexical access: missing word or jargon.

\item {\bf Syntactic troubles}

Most widespread, they appear in the subjects' communication as a more
or less flagrant destructuration of the utterances. The children reach
a stage in the development of syntactic competence and cannot progress
beyond that stage. This implies weak grammaticality, and frequently
goes along with subjects' preference to short utterances. Furthermore,
two noticeable trends have emerged from corpus analysis:

First, no, or very few, morphosyntactic information is inserted in the
message: absent or improper flexion, no ``grammatical words''. For
example, coordination is seldom explicitly conveyed by a particle,
neither are semantic relations like attribution, property~\dots The
subjects tend to use only ``meaningful'' words, producing
telegraphic-style utterances.

Second, the order of the words or symbols in the utterances is not
systematically determined by regular rules of grammatical nature. It
is chiefly guided by the focus of the message, leading to topicalized
utterances. Concepts do not go through a linear encoding of a deep
syntactic structure.

These observations led us to consider semantic analysis as the
appropriate way to get the meaning of these utterances.

\end{enumerate}

There are different symptom clusters in situations of dysphasia; but
the two main and most widespread symptoms are phonological (speech)
and syntactical (language) disorders. This study has been led with a
purpose of pragmatic communication aid, more than in a speech therapy
perspective. Hence we will focus on the syntactical disorders and the
methods proposed to make up for them.

\subsection{Adaptative and Augmentative Communication}

To make up for those difficulties in using language, rehabilitation
centers use a set of vicarious symbolic systems generically referred
to as AAC ({\em Adaptative and Augmentative
Communication})~\cite{kn:lloyd-quist-windsor}.

Several artificial languages have been developed for educational,
rehabilitation or communication purposes. These languages are grounded
on the preserved linguistic capabilities, which mainly consist in
loose categorization and semantic association. They do not rely on any
rigid structure, as syntax is beyond the reach of the language
impaired patients. These languages are symbolic or iconic and include
Bliss, Communimage and Grach.

\begin{enumerate}

\item The Bliss pictographic alphabet~\cite{kn:bliss} is composed of
ideograms which can be assembled with atomic ideographic elements. It
is the most elaborate of the three, and may represent some abstract
notions.

\item The Communimage icon set is composed of highly representative
figurative drawings.

\item The GRACH symbol set is also of an iconic nature, although more
stylized than the Communimage.

\end{enumerate}

Those languages offer:

\begin{enumerate}

\item an easier access to meaning, as many of these systems use
figurative icons. Even the Bliss alphabet is based on a non-arbitrary
relation between a symbol and its signified concept;

\item correlated to the previous point, a more limited set of symbols,
excluding in particular subtleties for abstract notions, and excluding
``empty'', i.e. grammatical words;

\item absence of a specified syntax, the iconic or pictographic
language offering simply a set of isolated symbolic conventions with
very few dialectal pressure (i.e. collectively set habits of using
them).

\end{enumerate}

The discourse that these symbolic systems allow the subjects to
produce is thus essentially based on semantics.

The utterances have an underlying semantic structure representing
their meaning, where the semantic units are linked to the others
through casual relations. This meaning is expressed by the mere
sequence of symbols corresponding to the semantic units, as there is
no way to express the type of casual relations. There is a
directionality in the semantic structure which is expressed by the
order of the symbols in the sequence.

While these iconic languages can be interpreted by medical staff, the
process of automating their interpretation through computer appears to
have several benefits, including giving a correct feedback on patients
(which can serve rehabilitation purposes) and enabling them to
communicate with a broader environment, not restricted to their family
and medical staff. Because these language have a finite set of
semantic contents, automatic processing also appears feasible.

\section{Semantic Analysis}

Therapists or parents of language impaired children generally
understand the children's messages because they reconstruct the global
meaning by attributing a correct semantic role to every word or
symbol. We tried to formalize this process so as to be able to
implement it in a communication help software.

The first step is to ground our work material on the phenomena
observed in the corpus. We collected a corpus reflecting the
spontaneous use of symbolic languages (mainly Bliss and Communimage)
by language-impaired children. This corpus is a set of icon sequences
(average length of four) which constitute single ``utterances'', each
one being usually interpreted by a Bliss-skilled nurse. Examples of
utterances in the corpus are: {\sf I/PUT/FLOWER/TABLE}, {\sf
I/WANT/SLEEP}, {\sf I/WANT/EAT/FISH/CAKE}, {\sf
ANIMAL/PLAY/BALL}~\dots

Study of the corpus led to consider two main semiotical facts:

\begin{enumerate}

\item paradigmatic structures: some sets of icons obviously form
semantic categories, as they may appear in the same contexts (e.g. the
category of ``{\em meals}'', which all come with the pictogram for
``{\em eating}'');

\item syntagmatic structures: some icons very systematically appear
along with some complemental icons, within the same sequence, which
belong to regularly the same categories (typically the pictogram
``{\em to eat}'' with an icon representing an animal or human being,
and with another icon representing a meal). {\em Syntagmatic}
structures don't mean {\em syntactic} structures, as no compulsory
order is always respected, but they form ``frames'' which represent
the basic context associated to a particular icon.

\end{enumerate}

These facts were to support a representation of the iconic language
which is exposed below:

\subsection{Cognitive and Linguistic Postulates}
\label{cognitive-postulates}

The language we are trying to analyse has the following two main
characteristics:

\begin{enumerate}

\item It is generated from a lexicon of invariant, meaningful words or
symbols.

Following linguistic evidence~\cite{kn:semantique-interpretative}, we
organized the lexicon into {\em ad hoc} categories arising from corpus
studies, {\em taxemes}. These taxemes are groups of symbols which have
a common semantic base and may be used in the same contexts, e.g. the
taxeme of beverages. Every taxeme is part of a {\em semantic
domain}. The domains give a frame for general semantic consistency of
the utterances.

The semantic content of a terminal in the lexicon is thus composed of:

\begin{enumerate}

\item a semantic domain;

\item a semantic category: the taxeme;

\item some specific semantic content distinguishing it from the other
members of the same taxeme.

\end{enumerate}

\item The utterances of the language are short sequences with no
formal structure where the main semantic units are disposed in a
topicalized order.

They have an underlying semantic structure representing their meaning,
where the semantic units are linked to the others through casual
relations. This meaning is expressed by the mere sequence of symbols
corresponding to the semantic units, as there is no way to express the
type of casual relations. There is a directionality in the semantic
structure which is expressed by the order of the symbols in the
sequence.

\end{enumerate}

It could be argued whether these postulates on the nature of the
language of dysphasic subjects are not oversimplifications of complex
disorders of the manipulation of syntactic structures. However we have
adopted them as a good approximation for short sequences of symbols.

Having thus pointed out the properties of the subjects' language,
postulated out of corpus evidence, we may define a model fit for
implementation.

\subsection{Formalization}

In order to manipulate the semantic content of the symbols, we use a
structural description based on semantic features. Every symbol has
generic features inherited from the domain and the taxeme it belongs
to, and specific features identifying it inside the taxeme.

A feature is defined as a simple attribute-value couple, where the
value is always an atom. In most cases in our lexicon, the elementary
features we use have a binary value: $+1$ or $-1$.

The number of features used to define the content of one symbol is not
set a priori, but depends on the needs to distinguish it from other
symbols. This approach, which is the approach of {\em differential
semantics}~\cite{kn:semantique-interpretative}, is based on the corpus
only, and ensures compatibility with assessed semiotic phenomena. It
has the drawback of setting combinatory problems when the size of the
lexicon grows, but we have been dealing up to now with a small corpus
and have not met the problem yet.

The meaning content of an utterance is represented by a network in
which the vortices are semantic units and typed arcs are casual
relations, like in Fig.\,1. Topicality is represented by an order
defined on the vortices of the network.

\vspace{2mm}

\noindent \makebox[60mm][c]{\parbox[t]{60mm}{\epsfysize=26mm
                            \epsffile{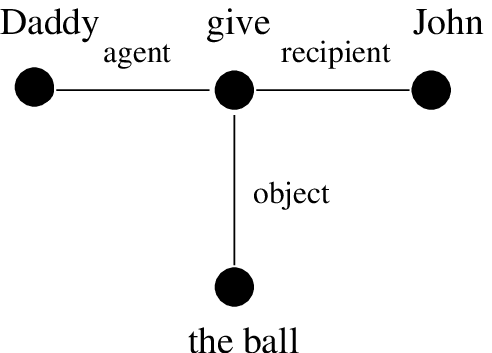}}}
          \makebox[60mm][c]{\parbox[t]{60mm}{\epsfysize=26mm
                            \epsffile{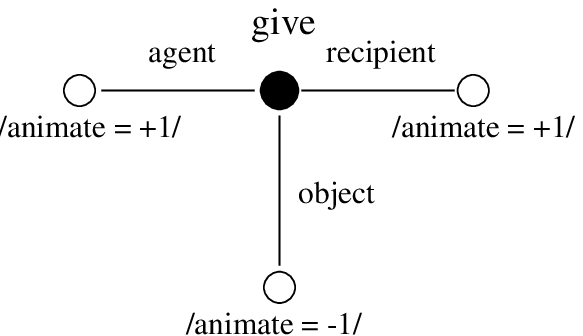}}}\\
\makebox[60mm][c]{\small {\bf Fig.\,1.}~The semantic network}
\makebox[60mm][c]{\small {\bf Fig.\,2.}~A potential casual structure}

\vspace{2mm}

The basic operation chosen to represent dynamic manipulation of
semantic data is {\em unification}~\cite{kn:kay}. A semantic relation
in a network is the actualization of a potential structure where some
variables are left uninstantiated. These potential structures are
typical casual structures, observable in the corpus, which are
``fossilized'' in the lexicon (like in Fig.\,2). The semantic
information borne by these structures is represented as selectional
features, which condition the unification of a symbol as the casual
filler of another.

\subsection{Heuristics for Automatic Understanding} \label{heuristics}

Natural Language Understanding systems are classically based on a
first step being the parsing of formal structures. \cite{kn:covington}
defines a dependency parser for free word order languages such as
latin, but it still relies on syntactic (to be exact, morphological)
information.

Our aim in this study is to provide a good analysis of a language
which has a limited expressive power, but provides no syntactic
information to guide understanding. As we have postulated
(\ref{cognitive-postulates}.2.) that this language is flatly generated
from a semantic network where the organization is provided by semantic
relations, we shall logically extract information from the sequences
by trying to identify these semantic relations, in order to find back
a semantic network.

This is done by trying to match the best case fillers to every
potential casual structure attached to a symbol in the sequence.

The input to analyse is a sequence of symbols $s_{1}, s_{2},
... s_{n}$, where every symbol in the sequence has a set of intrinsic
features defining its semantic content: ${\cal IF}(s_{i}) = F$ ($F$ is
a set of semantic features).

Some symbols in the sequence are ``predicative'' symbols, i.e. a
potential casual structure may be attached to them. The casual
structure is a set of casual relations, each of which has a set of
selectional features attached to it:

\[ {\cal CS}(s_{i}) = \{<c_{1},F_{1}>,<c_{2},F_{2}>,...<c_{k},F_{k}>\}
\]

(where $c_{j}$ is the type of a casual relation, and $F_{j}$ is a set
of semantic features).

We note the set of selectional features attached to the (predicative)
symbol $s_{i}$ for the case $c_{j}$: ${\cal SF}(s_{i},c_{j}) = F$
(which is equivalent to $<c_{j},F> \in {\cal CS}(s_{i})$).

We define the ``value'' of a case-filling unification, i.e. the value
of the symbol $s_{k}$ as a filler for the case $c_{j}$ of the
predicative symbol $s_{i}$, as the {\em semantic compatibility} of the
intrinsic features of $s_{k}$ to the selectional features of $s_{i}$
for the case $c_{j}$:

\begin{equation} {\cal V}(s_{i},c_{j},s_{k}) = {\cal C}({\cal
SF}(s_{i},c_{j}),{\cal IF}(s_{k})) \end{equation}

The relation of semantic compatibility of a set of semantic features
to another is itself defined as:

\begin{equation} {\cal C}(F_{1},F_{2}) = \frac{\sum_{f_{i} \in F_{1}
\cap F_{2}} {\cal X}(f_{i},F_{1},F_{2})}{\mbox{number of elements in
$F_{2}$}} \end{equation}

\noindent \begin{tabular}{lrcl} \makebox[2cm][l]{where} & ${\cal
X}(f_{i},F_{1},F_{2})$ & $=$ & $+1$ if $f_{i}$ has the same value in
$F_{1}$ as in $F_{2}$,\\
      & & $=$ & $-1$ otherwise.  \end{tabular}

\vspace{1mm}

This relation is asymetric: it measures the degree of fitness {\em of}
the set $F_{2}$ {\em to} the set $F_{1}$.

An {\em affectation} $A$ of a set of candidate symbols $S =
\{s_{i1},s_{i2},...s_{ij}\}$ as case-fillers to the predicative symbol
$s_{i}$ is an application of the set of cases of $s_{i}$ (${\cal
CS}(s_{i}) = \{<c_{1},F_{1}>,<c_{2},F_{2}>,...<c_{k},F_{k}>\}$) into
the set of candidate symbols:

\begin{equation}
  A = \{<c_{x},s_{iy}>\}\mbox{, where $x \in [1,k]$ and $y \in
[1,j]$.}  \end{equation}

We define the global value of an affectation $A$ of the symbols
$s_{i1},s_{i2},...s_{ik}$ as case-fillers of the predicative symbol
$s_{i}$ as the sum of the values of every single unification:

\begin{equation} {\cal
V}(s_{i},\{<c_{1},s_{i1}>,<c_{2},s_{i2}>,...<c_{k},s_{ik}>\}) =
\sum_{j \in [1,k]} {\cal V}(s_{i},c_{j},s_{ij}) \end{equation}

Hence, the search of a best interpretation of the sequence is the
search, for every predicative symbol of the sequence, of the best
affectation of other symbols as its case-fillers, i.e. the search of a
maximum for the value defined above:

\begin{equation} \begin{array}[t]{c}
  {\rm max} \\ {\scriptstyle A} \end{array} {\cal V}(s_{i},A)
\end{equation}

\subsection{Implementation}

Sample {\sc Prolog} code is provided to illustrate the implementation.

Intrinsic features of the symbols are defined in the internal
database:

\begin{verbatim} feature(Sym,(Att,Val)).  \end{verbatim}

So are the predicative symbols' selectional features, attached to
their casual relations:

\begin{verbatim} case(Sym,Cas,(Att,Val)).  \end{verbatim}

The semantic compatibility of a set of semantic features to another is
calculated based on the number of selectional features satisfied by
the presence of the corresponding intrinsic features (with the same
value):

\begin{verbatim} compatible_ratio(L,[],(0,Den)) :-
  length(L,Den).

compatible_ratio(L1,[(Att,Val)|L2],(Sum,Den)) :-
  member((Att,Val),L1),!,
  compatible_ratio(L1,L2,(Psum,Den)),
  Sum is Psum+1.

compatible_ratio(L1,[(Att,Val2)|L2],(Sum,Den)) :-
  member((Att,Val1),L1),!,
  Val1 =\= Val2,
  compatible_ratio(L1,L2,(Psum,Den)),
  Sum is Psum-1.

compatible_ratio(L1,[(_,_)|L2],(Sum,Den)) :-
  compatible_ratio(L1,L2,(Sum,Den)).  \end{verbatim}

\begin{verbatim} compatible_float(L1,L2,Real) :-
  compatible_ratio(L1,L2,(Sum,Den)),
  Real is Sum/Den.  \end{verbatim}

The semantic value of an affectation is the sum of the semantic values
of every single unification of a symbol to a case:

\begin{verbatim} affectation(Pred,[],_,0).

affectation(Pred,[Cas|Lc],[Sym|Ls],Score) :-
  bagof((SelAtt,SelVal),
        case(Pred,Cas,(SelAtt,SelVal)),
        LselFeat),
  bagof((IntAtt,IntVal),
        feature(Sym,(IntAtt,IntVal)),
        LintFeat),
  compatible_float(LselFeat,LintFeat,UnifScore),
  affectation(Pred,Lc,Ls,Pscore),
  Score is Pscore+UnifScore.  \end{verbatim}

The search of the best affectations is then the result of a {\em quick
sort} algorithm.

\subsection{Other Elements of the Analysis Process}

With the analysis technique described in \ref{heuristics}, there
potentially could be a correct interpretation of any sequence of
symbols, provided that the total number of casual relations in the
casual structures reaches the number of symbols in the sequence minus
one. As a matter of fact the search for a maximum always yields a
result, even if the maximum is negative.

Pragmatically, this is unrealistic and might lead to utter
nonsense. The data given in the corpus show that a minimal isosemy is
present in any utterance, guaranteeing its consistency.

We have thus introduced a first constant, the {\em acceptability
threshold}. Individual semantic unifications whose values do not
exceed this threshold are rejected.

Similarly, the topicality of the utterances makes it unlikely that
long distance semantic attachments exist between two symbols which are
not in a close vicinity in the sequence. This locality constraint
becomes relevant as soon as sequences are 4 or 5 symbols long. To take
it into account, we have defined a second constant, the {\em locality}
constant, which represents the fading of semantic relations with the
linear distance in the uttered sequences.

Practically, this constant will intervene in the calculus of the value
of a semantic unification at the power of $n$, $n$ being the distance
between the two semantic units within the sequence.

This constant is a rough way of modeling the effect of distance in
semantic relations inside a text (in a broad sense). We use it
successfully on our small examples.

Both the acceptation threshold and the locality constant have been
defined by iterative tries based on the corpus.

\section{The Application}

The technique of semantic analysis described above has been
implemented in an adaptative and augmentative communication
application: PVI ({\em Proth\`{e}se Vocale Intelligente},
i.e. Intelligent Voice Prosthesis), available as a software program
for portable computers. This application has a broader scope which
also includes assisting the subjects for pictogram input, taking into
account their motor disabilities, as well as generating correct
sentences in natural language (French) from the semantic networks
obtained after the analysis.

As the differential semantic description appears to be common both to
symbolic languages and natural language, it provides the basis for
conversion of one language into another. To convert a semantic
representation into natural language assumes that the process of
semantic analysis can be somehow reversed, a processing phase called
lexical choice. It consists in determining which words can be formed
from the network of semantic features yielded by the semantic
analysis. This is mainly a matter of reorganizing the semantic content
into relevance islands. These islands are determined by the proper
description of an object or an action. For instance, every feature
describing the same object will be grouped into a single word - if
such a word exists - no matter which icon they come from. This is
performed through a natural language dictionary described with the
same semantic features as the icon vocabulary, and a set of
heuristics.

Of course, in natural language, even simple utterances have to follow
syntactic well-formdness principles. This is why conversion between
symbolic and natural language cannot rely purely on semantic knowledge
but has to include syntactic information in the late stage of
translation. Syntactic information is incorporated into syntactic
trees in the formalism of Tree-Adjoining Grammar~\cite{kn:joshi}. This
accounts for a predicate-centered syntactic representation accepting
various modifiers which fit the basic phenomena encountered. More
complex syntactic phenomena such as long-distance dependencies fall
out of our scope.

As a whole, the PVI application should be a completely transparent
application with a customizable, graphical front-end for the user, and
a natural language front-end for the interlocutors, ideally acting as
a filter between agrammatical pictographic designation and natural
language.

Sample utterances treated by the application:

\vspace{1mm}

\noindent \parbox[b]{78mm}{ \verb+ ?-
transfer([i,eat,meat],Sentence).+\\ \verb+ Sentence = "Je mange la
viande" + } \hfill {\em ``I eat the meat''}

\vspace{1mm}

\noindent \parbox[b]{78mm}{ \verb+ ?-
transfer([meat,i,eat],Sentence).+\\ \verb+ Sentence = "Je mange la
viande" + } \hfill {\em ``I eat the meat''}

\vspace{1mm}

\noindent \parbox[b]{78mm}{ \verb+ ?-
transfer([fork,i,eat],Sentence).+\\ \verb+ Sentence = "Je mange avec
la fourchette" + } \hfill {\em ``I eat with the fork''}

\vspace{1mm}

\noindent \parbox[b]{76mm}{ \verb+ ?-
transfer([fork,i,eat,meat],Sentence).+\\ \verb+ Sentence = "Je mange
la viande+\\ \verb+ avec la fourchette" + } \hfill {\em ``I eat the
meat with the fork''}

\vspace{1mm}

\noindent \parbox[b]{76mm}{ \verb+ ?-
transfer([i,give,cat,meat],Sentence).+\\ \verb+ Sentence = "Je donne
la viande au chat" + } \hfill {\em ``I give the meat to the cat''}

\vspace{1mm}

\noindent \parbox[b]{78mm}{ \verb+ ?-
transfer([i,give,cat,daddy],Sentence).+\\ \verb+ Sentence = "Je donne
le chat à Papa" + } \hfill {\em ``I give the cat to Daddy''}

\section{Evaluation}

The system has been submitted to a benchmarking test: a set of 200
icon sequences, reproducing in their structure a number of
spontaneously uttered patterns, has been given as input for content
analysis and language generation.

The results were indexed into the four following categories, depending
on their correct analysis but also on the ``naturalness'' of the
French sentence produced: (I)~Correct analysis, correct generation;
(II)~Correct analysis, clumsy generation; (III)~Incomplete or clumsy
analysis; (IV)~Incorrect analysis.

The results were the following: category I: 147 sequences; category
II: 15 sequences; category III: 15 sequences; category IV: 18
sequences.

When we decide to consider ``acceptable'' the sequences which were
either correctly analysed and generated, or correctly analysed but
imperfectly generated (and still comprehensible), that is when we
merge categories I and II, we thus get an acceptability rate of
80.5\,\% on this benchmark.

This of course must not be taken for a global acceptability level of
the PVI system by the user in an ecological situation. During on-site
evaluation, which was conducted during five months with six
individuals subject to cerebral palsy in the rehabilitation center of
Kerpape (Brittany, France), a certain number of problems linked to
real-life situations were unveiled:

\begin{enumerate}

\item unexpected answers from the system, even if they are a minority,
very soon get the user frustrated and nervous, since the actual input
of the sequence of icons by a person suffering from motor disabilities
is rather long (it may be counted in minutes). A bad result is thus
immediately resented as a frustrating waste of time;

\item lack of vocabulary can not easily be overriden by hand gestures
or segments of words, as it is the case during direct
communication~--- or else the missing element in the sequence will
lead to nonsense. We have been asked with emphasis to increase the
initial vocabulary of the system (grossly 300 icons) whose limits are
reached very soon;

\item problems of interface ergonomy are sometimes crucially important
for users who have only a few interface points with the system.

\end{enumerate}

However, these critiques might be interpreted as an encouragement to
develop a promising prototype, whose principles have been validated,
and to adapt it to the realities of difficult ecological situations.

\section{Conclusion}

We have proposed and implemented a semantic analyser which performs
manipulation of symbolic knowledge. It has proved to be successful in
the interpretation of weakly structured utterances of symbols.

Further reflexion will aim at taking into account other semantic
phenomena. Very interesting ones are contextual meaning effects, which
should be described by {\em dynamic} manipulation of semantic
features.

Another arising topic of interest is the more detailed theoretical
study of how visual (icon or pictogram) semantics and language
semantics intersect and interact wih each other.

The technique exposed in this paper was designed to cope with a
specific problem of language alteration for dysphasic subjects, but
its availability might open new perspectives. Language prostheses have
a great potential for the rehabilitation of language impaired
patients. In particular, its adaptation to adult traumatic aphasia,
with the experience in the field of rehabilitation for these cases,
might bring promising results.

\end{document}